\documentclass[runningheads]{llncs}
\usepackage[T1]{fontenc}
\usepackage{graphicx}
\usepackage{amsmath}
\usepackage{booktabs} 
\begin{document}
\title{Efficient Multi-Band Temporal Video Filter for Reducing Human-Robot Interaction}
%
%
\author{Lawrence O'Gorman}
\institute{Nokia Bell Labs, Murray Hill, NJ, USA\\
\email{larry.o\_gorman@nokia-bell-labs.com}}

\maketitle              
\begin{abstract}
Although mobile robots have on-board sensors to perform navigation, their efficiency in completing paths can be enhanced by planning to avoid human interaction. Infrastructure cameras can capture human activity continuously for the purpose of compiling activity analytics to choose efficient times and routes. We describe a cascade temporal filtering method to efficiently extract short- and long-term activity in two time dimensions, isochronal and chronological, for use in global path planning and local navigation respectively. The temporal filter has application either independently, or, if object recognition is also required, it can be used as a pre-filter to perform activity-gating of the more computationally expensive neural network processing. For a testbed 32-camera network, we show how this hybrid approach can achieve over 8 times improvement in frames per second throughput and 6.5 times reduction of system power use. We also show how the cost map of static objects in the ROS robot software development framework is augmented with dynamic regions determined from the temporal filter.

\end{abstract}
\section{Introduction}
\label{sec:Introduction}

Robots must navigate with respect to both their static world (walls and fixed objects) and dynamic (people and other robots). The dynamic world can be classified in terms of short- and long-term time frames. Robots capture short-term events by using their on-board sensors; for instance, a person steps in front of the robot and the robot should stop. But there is also activity that repeats in predictable, longer-term periodic cycles. Repetition over a regular time frame is termed \textit{isochronal}.  Examples of isochronal time periods include factory shifts, scheduled deliveries, and employee breaks. The effects of these activities on navigation are just as real as for static objects except that their occurrence is time dependent. In this paper, we determine both long- and short-term activity by temporal video filtering for use in robot navigation and path planning. 

Video analysis of human activity can be performed using convolutional neural networks or vision transformers to detect and track people. This neural network processing can achieve a high level of recognition, but at high cost of computation. If only activity detection is required -- not individually segmented persons -- then motion flow \cite{Grimson2009,Nishino2009,Shah2010,motionOGorman2014} is a less expensive alternative. Instead of GPU-processing needed for real-time neural network detection, motion flow can be performed by IoT-level processors, typified by low cost, low power, small memory, and narrow bandwidth. This low level of processing is a practical cost- and power-usage alternative for installations that might have tens or hundreds of cameras. A hybrid solution including a temporal filter and neural network object detection is also shown to realize cost and power efficiencies.

The application goal of this paper is to use fixed cameras to detect human activity such that it can be avoided for the purpose of efficient robot path planning and navigation. We consider both off-line global path planning, where the goal is to schedule the robot for regular (daily, etc.) tasks on selected paths and at times that are efficient and safe with respect to human activity; and real-time local navigation where the goal is to choose the best of current path options for immediate robot navigation.

The technology goal of this paper is to offer an efficient multi-band temporal video filter for extracting short- and long-term activity bands from both chronological and isochronal time. Extraction of these bands requires multiple low, high, and bandpass filters. These could be implemented separately, however we show how a cascade filter architecture can extract all these bands efficiently from a single video stream. We show how use of the cascade filter both reduces video processing and video storage. Although temporal video filters are common and activity detection to avoid human-robot interaction is often used, we believe the design and use of a single cascade filter to efficiently extract multiple bands over chronological and isochronal time is novel.

The main contributions of this work are:
\begin{enumerate}
\item A cascade filter that extracts temporal video information of long- and short-term human activity more efficiently than through separate filters.
\item Use of a single, efficient cascade filter to identify long- and short-term activity to aid global and local robot navigation.
\item An efficiency analysis of using pixel- and feature-based activity analytics either independently or as a hybrid combination of pre-filter and neural network object detection.
\item Practical implementation on the ROS robot operating system.
\end{enumerate}

In Section \ref{sec:Background}, we review related literature . In Section \ref{sec:Method}, we describe the system architecture to extract long- and short-term activity. Section \ref{sec:Exp} shows costs of computation of activity detection, object detection, and a hybrid of both.

\section{Background}
\label{sec:Background}
Early work in robot navigation dealt with a static environment of building walls and fixed-placed objects, e.g., \cite{Kavraki1996}. Inclusion of moving objects (other robots) in dynamic environments followed, e.g., \cite{Belkhouche2009}. While navigation with respect to inanimate objects is a challenge, human presence adds the trade-off of safety versus efficiency. In \cite{Wada2010} the procedure of SLAM (Simultaneous Localization and Mapping) is augmented to include humans. For prolonged observation, the SLAM robot can both observe and extrapolate human trajectories to create human motion maps. However, there is a degree of unpredictability when dealing with humans \cite{Charalampous2017} that makes the success of trajectory prediction variable.

Because full trajectories are often difficult to track, many approaches represent floor space as an occupancy grid and determine statistics independently within each grid cell \cite{Moravec1985}. In \cite{Saarinen2012}, this is done with 2-state Markov probabilities of entry and exit to a cell. Direction is added in \cite{Wang2016}, in which a 9-state Hidden Markov Model describes motion direction from each grid cell, and a 9th state for staying in the same cell. In \cite{Jumel2017}, grid flow is extended to be either observed (statistical) or spatially extrapolated from cell directions to predict continuing trajectories. Mobile robots cannot be in all places at all times so it is understandable that predicted flow is a valuable complement to observed flow. Finally, work such as \cite{Wu2019} combine methods discussed here to yield a multi-layer representation (static layer from SLAM and object layer from YOLO \cite{Redmon2016}).

Besides avoiding human-occupied areas, advantage can be gained by observing the paths humans travel and to follow these. Imitation learning, or inverse reinforcement learning, is a machine learning approach that doesn't require training with labeled samples. Instead, an agent observes how experts behave (humans in our case), learns a reward function that the experts are unconsciously acting upon, and seeks to maximize that reward \cite{ElShamoutyReinfLearn2020}. In \cite{Ziebart2009} human trajectories are observed to learn their normal paths with respect to objects. With this prediction, efficient human-aware robot paths can be planned. In \cite{Kuderer2013}, inverse reinforcement learning is used with particular emphasis on socially normative navigation in dense and complex scenes such as meeting places and hallway intersections. This can be extended beyond just navigation to where robots can learn more complex human movements for the purpose of human-robot collaboration \cite{Collaboration2022}. 

Besides static location of objects and prediction of forward path, affordance is another relevant factor for robot navigation among objects and humans. Affordance describes how an object is used, and for navigation purposes this relates to spatial interaction between human and object \cite{Koppula2016,Truong2017,Affordance2023}.

For the previously described work and for many robot navigation systems, sensors on the robot are used for navigation. But many situations limit robots to indoors and on paths traveled repeatedly. In these cases, fixed cameras can augment onboard sensors to aid navigation. In \cite{Ravankar2020}, fixed cameras are used to create a heat map-based path planner. Motion pixels are found and accumulated into “heat values”. Resulting cost values at regular grid locations are associated with their closest path edges. This reduction from grid points to many fewer path edges reduces storage and subsequent communication of cost values to the robot. A relatively new fixed-camera alternative is an event camera, which contains bio-inspired vision sensors to capture scene changes \cite{IaboniEventCamera2021}. Although these capture activity, as is our goal, they do not also capture traditional frames for video processing as is also our goal, so are outside the domain of this paper.

This paper has similarities and differences with respect to the literature described. Unlike work that combines SLAM and person detection \cite{Moravec1985}, we detect only people, but do so by their activity rather than their identity. Unlike work that categorizes objects by their affordances \cite{Koppula2016}, we deal only indirectly by learning observed human activity and creating a model similar to cost maps and social force models \cite{Kuderer2013,Okal2016,Truong2017,Ravankar2020}. A difference in our work from cost maps and social force models, which directs a robot \textit{away from} obstacles is that our model directs it \textit{toward} higher probability paths. In contrast to work using on-board robot sensors and cameras to aid navigation \cite{Belkhouche2009,Wada2010,Moravec1985,Jumel2017,Wu2019,Okal2016,Truong2017}, we use fixed cameras as do \cite{Wang2016,Ziebart2009,Ravankar2020}. There is much work in learning and avoiding humans in close-up human-robot interaction with static robots \cite{Kumar2021,Arzani2021}, which has both similarities and differences to mobile robot interaction investigated here.

Our work is closest in purpose and methods to those proposing dynamic occupancy grids \cite{Saarinen2012,Wang2016,Jumel2017,Ravankar2020}. These are created by unsupervised learning of human activity in a grid-space over time. Our work also has similarities to the inverse reinforcement learning approaches used for predicting trajectories in \cite{Ziebart2009} and socially normative behaviors in \cite{Kuderer2013,Okal2016}. Whereas these seek to generalize beyond specific objects and locations, our approach imitates what humans do with emphasis on fixed locations and times. 

\section{Method}
\label{sec:Method}

\subsection{Definitions}
\label{sec:Definitions}
Our methods distinguish different types of human activity related to different navigation tasks as shown in Table \ref{tbl:planVsActivity}. \textit{Long-term activity} refers to human motion that is statistically stationary in time and place. We also use the term isochronal, meaning that this activity happens on a cyclic basis in some time frame. For simplicity in this paper, our time cycle is one day, so long-term activity refers to the activity that is statistically determined over many days at each chosen time of day. We designate isochronal time as $t^*$, so an example of an isochronal sequence is {${t_1}^* = 18:23$ Monday, ${t_2}^* = 18:23$ Tuesday, …}.

 \textit{Short-term activity} refers to human motion at the current time, of which we distinguish two types. \textit{In-place activity} is static in location. In-place activity may include people who are stationary in location such as waiting in line or dwelling at a shop window. In-place activity also includes people who are not stationary in location, but who create a location that is active by, for instance, passing through a crowded bottleneck such as an entranceway. Opposite to in-place activity is \textit{moving activity}. This refers to people movement with changing location, such as people walking.

\begin{table}[t]
\centering
   \caption{Activity types corresponding to path planning and navigation.}
\setlength{\tabcolsep}{8pt} 
\begin{tabular}{c c c c c}
\toprule
Activity & Global & Global & Local & Navigation \\
 & Planning & Planning & Planning &  \\
 & Off-line & Real-time &  &  \\
\midrule
Long-term & 1 & 2 & -- & --  \\
Short-term, in-place & -- & 1 & 1 & 1  \\
Short-term, moving & -- & 1 & 1 & 1 \\
\end{tabular}
   \label{tbl:planVsActivity}
\end{table}

 We distinguish two types of global path planning. \textit{Off-line global path planning} pertains to the task of choosing a robot’s full path for a future time. If we are arranging a planned daily trip of a robot delivery cart for example, we would seek to choose the times and paths that are statistically of least activity. \textit{Real-time global path planning} pertains to planning a full path that is to be begun at the current time. \textit{Local planning} pertains to altering the global path with information local to (i.e., a short distance from) the robot at that time.

In Table \ref{tbl:planVsActivity}, off-line global path planning can only be performed with respect to long-term activity because short-term activity is not known off-line. However, for real-time global path planning, in-place short-term activity can be used because both are happening at the current time. We designate in-place short-term activity as first choice ``1'' for this column and second choice ``2'' for long-term activity. This is because, when a path is altered due to short-term activity, the real-time global planning may also use long-term activity information.

\subsection{Architectural Overview}
\label{sec:Arch_Overview}

\begin{figure}[t]
\centering
\vspace{0.4cm}
    \includegraphics[width=0.9\linewidth]{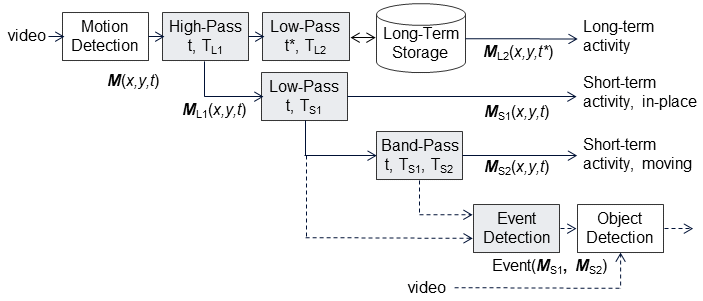}
   \caption{Functional diagram shows temporal filter cascade and event detection.}
   \label{fig:functionalDiag}
\end{figure}

A functional diagram of the cascade filter is shown in Fig. \ref{fig:functionalDiag}. Motion detection is performed on each video frame, then a cascade of temporal video filters extracts long- and short-term activity. We describe each component in sections below. 
 
\subsection{Motion Detection}
\label{sec:MotionDetection}

Motion detection is performed on each frame to obtain a motion image of $K$ blocks subsampled from the full frame, each block $k$ containing 2 motion features, $\textbf{f}_k =$ (density, direction), corresponding to an $(x,y)$ location,
\begin{equation}
\textbf{M}(x,y,t)=\{f_k \}_t, 0 < k < K
\label{eqn:motionFrame}
\end{equation}

Density is a measure of the motion in a block, a function of the number of motion pixels and their gradient values. Direction is quantized to 8 angles. The motion features are found from motion flow \cite{motionOGorman2014} or optical flow \cite{Fortun2015} methods. For brevity below we write a single block as $b(t)$, where $\textbf{b}(t)=\textbf{M}(x_i, y_j, t)$.

\subsection{Temporal Filtering}
\label{sec:TemporalFiltering}
\begin{figure}[t]
\centering
    \includegraphics[width=0.8\linewidth]{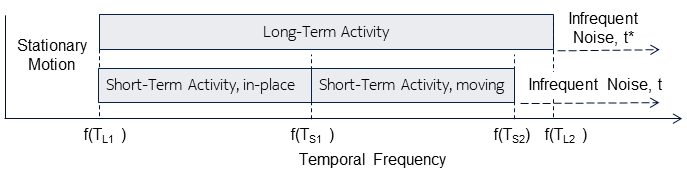}
   \caption{Temporal filter bands associated with activity types.}
   \label{fig:fltrBands}
\end{figure}

After motion detection, long- and short-term activity are found using a cascade of temporal video filters shown in Fig. \ref{fig:functionalDiag}. Fig. \ref{fig:fltrBands} shows the temporal filter bands of frequencies corresponding to the filter time constants TL1, TL2, TS1, and TS2, which are described below.

For all filtering (with one exception noted below), we use a first-order IIR filter, also called an exponential moving average filter, to give more weight to the most recent block $\textbf{b}(t)$ than past blocks $\textbf{b}^\prime (t - 1)$, and obtain the block result $\textbf{b}^\prime (t)$,
\begin{equation}
\textbf{b}^\prime (t)=\alpha \textbf{b}^\prime (t-1)+(1-\alpha)\textbf{b}(t), \textrm{   }\alpha \in [0,1]
\label{eqn:bPrimeT}
\end{equation}
	
 We choose the filter parameter value $\alpha$ through a more intuitive parameter, which we call the 10\%-decay duration, T. This is the amount of time during which a filtered signal will decay to 10\% of original with zero input. In equation \ref{eqn:bPrimeT}, if input $\textbf{b}(t)=0$ for $n$ samples, then $\textbf{b}^\prime (t=n)/\textbf{b}^\prime (t=0) = \alpha^ n = 0.1$. So, we can obtain $\alpha$ with chosen $T$ as follows,
\begin{equation}
\alpha =0.1^{(1/n)},n=rT,
\label{eqn:alpha}
\end{equation}
where the number of samples is equal to the video frame rate $r$ in frames per second times the 10\%-decay duration $T$ [sec]. The temporal video filters are described in more depth in \cite{OGorman2018}. 

Filtering begins in Fig. \ref{fig:functionalDiag} with a high-pass filter $F_{L1}$ applied to the frame-rate stream of motion vectors,
\begin{equation}
\textbf{M}(x,y,t) * F_{L1} \rightarrow \textbf{M}_{L1} (x,y,t)
\label{eqn:fltrFL1}
\end{equation}
The filter time constant $T_{L1}$ is chosen to reduce low frequency “stationary motion noise” as described in Section \ref{sec:fltrParams}.

The result of equation \ref{eqn:fltrFL1} is combined with the long-term, isochronal motion vector from storage at corresponding time $t=t^*$ using a low-pass filter $F_{L2}$, and the resultant $\textbf{M}_{L2}$ is stored,
\begin{equation}
\textbf{M}_{L1} (x,y,t^* )*F_{L2} \rightarrow \textbf{M}_{L2} (x,y,t^*)
\label{eqn:fltrFL2}
\end{equation}

Long-term activity is updated in isochronal time, in our case 1 sample per day for each $t^*$, $0<t^*<1440$, where 1440 is the number of minutes in a day. Because of this long sample period, there is a tradeoff between the duration of samples needed to obtain a good measure of long-term activity at any $t^*$ and the delay within which the measure adapts to changes in long-term activity. We choose a low-pass filter value to reduce infrequent (shot) noise as described in Section \ref{sec:fltrParams}.

Short-term, in-place activity can be identified by applying a low-pass filter $F_{S1}$ to $\textbf{M}_{L1}$, 
\begin{equation}
\textbf{M}_{L1} (x,y,t) * F_{S1} \rightarrow \textbf{M}_{S1} (x,y,t)
\label{eqn:motion_ML1}
\end{equation}
The time constant $T_{S1}$ is set to capture people activity in the same location, as described in Section \ref{sec:fltrParams}. 

Short-term, moving activity is identified using a band-pass filter. The low end of the filter is $T_{S1}$ and the high end $T_{S2}$. The time constant $T_{S1}$ separates the signal from in-place activity and $T_{S2}$ removes high-frequency, infrequent noise. Short-term, moving activity is found by subtracting $\textbf{M}_{S1}$ from $\textbf{M}_{L1}$ to rid the in-place activity (effectively a high-pass filter) and then applying a low-pass filter to rid infrequent noise. This combination results in band-pass filtering,
\begin{equation}
(\textbf{M}_{L1} (x,y,t) - \textbf{M}_{S1} (x,y,t)) * F_{S2} \rightarrow \textbf{M}_{S2} (x,y,t)
\label{eqn:BPFltr}
\end{equation}
Note that the two filters at the high frequency end of Fig. \ref{fig:fltrBands} are not redundant because one reduces noise in isochronal time $t^*$ at $T_{L2}$, and the other in chronological time $t$ at $T_{S2}$.

Finally, event detection is performed on short-term, in-place and moving activity to act as a gate on more computationally expensive processing such as object detection,
\begin{equation}
\begin{split}
\textrm{Event}(\textbf{M}_{S1},\textbf{M}_{S2}) &=1,\textrm{do object detection}\\
                                  &= 0, \textrm{do nothing}.
\end{split}
\label{eqn:eventDetect}
\end{equation}

\subsection{Off-line Global Path Planning}
\label{sec:OffLineGlobal}

\begin{figure}[t]
\centering
    \includegraphics[width=0.8\linewidth]{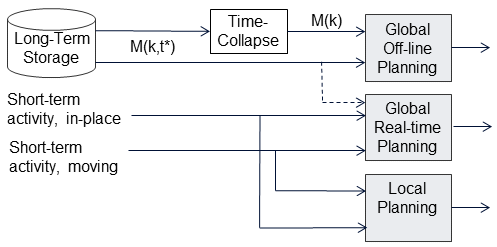}
   \caption{Different types of activity information used for different path planning tasks.}
   \label{fig:pathPlanOptions}
\end{figure}

Most commonly, a path is planned that avoids human activity in time and space. It is less common to choose path segments of high activity, but we do this in the following way. In Fig. \ref{fig:pathPlanOptions}, the long-term storage contains motion statistics for each minute of the day $\textbf{M}_p (f,t^*)$, where subscript $p$ indicates this is from a camera viewing path segment $p$. We time-collapse and binarize this as follows, 
\begin{equation}
\begin{split}
\textbf{M}_p  (f)&=1,\textrm{  if  } f(t^* ) \neq 0 \textrm{     for any } t^*, 0< t^* <1440\\
                                  &= 0, \textrm{ otherwise}.
\end{split}
\label{eqn:Mp}
\end{equation}

The result ${\textbf{M}^ \prime }_p$ can be thought of as a location attribute learned from people activity. If ${\textbf{M}^\prime}_p$ is 0, then no people travel through this location for whatever reason, so it’s reasonable for a robot to avoid it as well.

Off-line path planning through any location at $t^*$ is now a function of two activity-related values, $\textbf{M}_p(t^*)$ and $\textbf{M}^\prime _p$. The activity-related cost for a potential path through locations $p_i$ for off-line planning can be written,

\begin{equation}
\begin{split}
\textrm{Cost}_1 (\{ p_i \},t^ *)=& \sum_i \textrm{Cost} (\textbf{M}_{p_i}(t^*)), \textrm{   if all }{\textbf{M}^\prime}_{p_i} = 1\\
                =& \infty,    \textrm{             if any } {\textbf{M}^\prime}_{p_i} = 0.
\end{split}
\label{eqn:cost1}
\end{equation}

\subsection{Real-time Global Path Planning}
\label{sec:RealTimeGlobal}

Since real-time global path planning is performed just before the robot begins a path, there is current short-term activity information available as well as long-term information as shown in Table \ref{tbl:planVsActivity}. It may make sense to weight the short-term information higher than the long-term information $\{w1, w2\}$, although we do not explore that further here. For simplicity, we do not repeat the second line of equation \ref{eqn:cost1}, leaving it implicit that any locations of ${\textbf{M}^ \prime}_p =0$ are not included in a path. The activity-related cost for a path through segments $\{p_i\}$ from off-line planning is,
\begin{equation}
\begin{split}
\textrm{Cost}_2 (\{ p_i \},t))=&w_1 \textrm{Cost}_1 (\{p_i \},t^*=t)\\
                        &+w_2 \sum_{i=1}^N \textrm{Cost}(\textbf{M}_{p_i} (t)).
\end{split}
\label{eqn:cost2}
\end{equation}
The top line of equation \ref{eqn:cost2} is the long-term activity cost at isochronal time $t^ *=t$, which for real-time planning is the current time of path planning. The cost in the bottom line includes both in-place and static short-term activity. Short-term moving activity in Table \ref{tbl:planVsActivity}, which is captured from the on-board robot sensors for local planning, is outside the focus of this paper, but if captured it would be added to equation \ref{eqn:cost2}.

\section{Experiments and Results}
\label{sec:Exp}

\subsection{Scope of Experiments}
\label{sec:scope}
The focus of this paper is on efficient design of an activity filter and application to human-robot interaction. It is important to state what is outside the scope of this paper. Experiments showing the effectiveness of activity filtering with the same filters but not the same efficient architecture have already been described in \cite{motionOGorman2014}. Other references describe the performance of activity filtering on a variety of datasets and applications \cite{crowdReview2021}. We do not repeat these. This paper is also not a comparison between pixel-based activity filtering and neural network object detection. The former only detects activity; the latter can detect activity as well but in addition detect higher level features. However, we do show how their hybrid combination can yield both levels of information in an efficient manner.

\subsection{Filter Parameters}
\label{sec:fltrParams}
Filter parameter values are determined by balancing the signal-to-noise ratio for noise conditions learned statistically for each particular deployment. The values described in this paper are for our deployment of robot path planning in a factory setting described in Section \ref{sec:DeploymentROS}. 

The filter $F_{L1}$ in equation \ref{eqn:fltrFL1} is designed to reduce ``stationary motion noise''. This is motion that occurs in-place and continuously such as from rustling tree leaves or a flashing light. Choice of the value has a wide tolerance, the main consideration being that it should not be too short to reduce activity of interest. We choose to remove motion of duration 30 minutes and longer, so at 30 frames per second, $r=30$, $T_{L1} = 30\times 60$, and equation \ref{eqn:alpha} yields $\alpha=0.794$. 

For filter $F_{L2}$ in equation \ref{eqn:fltrFL2}, we choose a low-pass filter to reduce infrequent (shot) noise. To accomplish this, we choose a filtering duration of 10 days (this is 10 samples in isochronal time), so at 1 frame per day, $r = 1$, $T_{L2} = 1 \times 10$, and equation \ref{eqn:alpha} yields $\alpha=0.999957$.

For filter $F_{S1}$ in equation \ref{eqn:motion_ML1}, we choose a low-pass filter to capture people activity in the same location and eliminate people moving across locations. The discrimination between static and moving activity is somewhat arbitrary, so the filter value choice also has tolerance. We choose a low-pass filter with time constant greater than or equal to 20 seconds to define this activity, and this activity is updated not at frame rate but at 1/sec, so $r=1$, $T_{S1}=20$, and $\alpha =0.89$.

The band-pass filter of equation \ref{eqn:BPFltr} uses $F_{S1}$ on the low end, which is already specified. On the high end, $F_{S2}$, it is set to remove high-frequency, infrequent noise. This is the most intolerant of the filter parameters, since this noise has variable periodicity. We use a FIR filter to average activity values over 1 second. 

Finally, event detection is based upon the filtered results exceeding the activity average (Fig. \ref{fig:activityPlots}) plus chosen standard deviation. There are 2 components of activity, density and direction (equation \ref{eqn:motionFrame}). This number is far fewer than the 100s of features learned in a neural network. So for a hybrid solution, the standard deviation of the filter output can be chosen conservatively to allow more events, and depend upon the final stage neural network to reduce false events. 

\subsection{Isochronal Activity}
\label{IsochronalAct}

\begin{figure}[t]
\centering
    \includegraphics[width=1.0\linewidth]{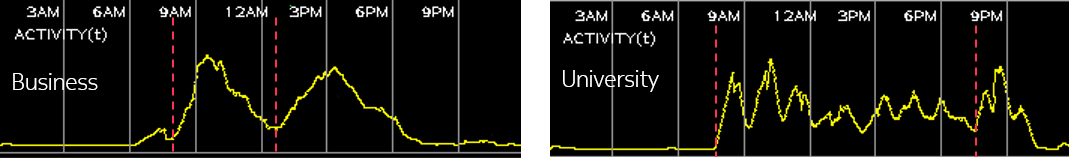}
   \caption{Isochronal activity plots showing magnitude of hallway activity in a business place (left) and a university (right).}
   \label{fig:activityPlots}
\end{figure}

Fig. \ref{fig:activityPlots} shows two examples of isochronal activity with 1-day periodicity. The top plot is from an office hallway. The data was collected and averaged over 2 years. It shows an increase of activity starting at 6am, a lull in mid-day, increase to 4:30pm, and activity decreasing to 9pm. The lower plot is data from a university hallway showing activity collected and averaged over 1 month of the school term. It shows rises and falls coinciding with hourly class changes. The red vertical lines show times where robot navigation might best be planned (within the work or school day) to avoid high activity periods.

\subsection{Cascade and Non-Cascade Filter}
\label{sec:CascadeVsNon}

For our application of using fixed cameras to monitor human and robot spaces, an industrial or business installation may use hundreds of cameras. It is important to limit costs of hardware and computation. We compare the computation and memory cost of the cascade temporal filter described in Section \ref{sec:TemporalFiltering} and Fig. \ref{fig:functionalDiag} against a non-cascaded filter that accomplishes the same task.

In Fig. \ref{fig:functionalDiag}, there are 5 filters, where the bandpass filter counts as 2 filters, a combined low- and high-pass filter. So a non-cascaded filter requires 5 filter operations. A cascade filter economizes by using the low-pass filter ($F_{S1}$) for both the short-term in-place and moving activity operations. Therefore, the cascade filter has an advantage of 4 versus 5 filtering operations.

For memory, both cascade and non-cascade filters require the long-term storage. For sequential filtering, the cascade filter needs only one storage for both short-term in-place and moving activity. However, the non-cascade filter needs storage for both. Therefore, the cascade filter has an advantage of 2 versus 3 motion feature frames. Results of this comparison are shown in Table \ref{tab:CompareFltrs}, in which \textit{Multiplies} is a multiple of motion frame filter operations and \textit{Memory} is a multiple of motion frame size.

 \begin{table}[t]{
 \begin{center}
    \caption{Computational costs of non-cascade versus cascade filter.}
    \setlength{\tabcolsep}{8pt} 
\begin{tabular}{c c c c}
\toprule
Cost & Non-Cascade & Cascade & Reduction \\
\midrule
Multiplies & $5\times$ & $4\times$ & 20\%  \\
Memory & $3\times$ & $2\times$ & 33\%  \\
\end{tabular}
\label{tab:CompareFltrs}
     
 \end{center}
}
\end{table}

\subsection{Cost of Computation of Activity and Object Detection}
\label{sec:Comparing}

Due to the high accuracy of neural network object detection (which we subsequently shorten to \textit{object detection}), this is likely to be the first choice of many practitioners for detecting humans to reduce human-robot interaction. This will indeed perform the task well, but at a relatively high computational cost. In this section, we show the computational cost of activity and object detection, and in the following section of a hybrid of both. 

Activity detection and object detection are different operations, the latter being much more versatile than the former. By extracting information on number of people, their pose, etc., an object detector can extract much more reliable information than a temporal filter. Our comparison in this section is strictly computational of combinations of solutions. Where low cost of computation is important, the more lightweight activity filter may be all that is needed. Where higher-level information is needed, the activity filter can act as pre-filter to an object detector to form a hybrid solution to reduce overall computation. And where the reliability or additional information of an object detector is always needed, we compare these costs as well.

Table \ref{tbl:benchmarkTable} shows computation results of comparing different methods for activity detection. Object detection methods are included in publication order, Faster R-CNN \cite{Ren2015}, YOLOv3, and Tiny Yolo \cite{Redmon2016}. We have added OpenPose \cite{Cao2019}, which finds people as well as their poses, because pose can be useful when working with affordances (as described in Section \ref{sec:Background}). The computational requirements were measured from a testbed 32-camera network viewing hallways and public areas of a building containing offices and laboratories. The computing specifications for processing the video streams are, CPU: AMD Ryzen 5 Pro 2600, 6-core, 8GB RAM; and GPU: NVIDIA GeForce GTX 1060, 6GB.

\begin{table}[t]
\centering
   \caption{Comparing methods for detecting activity.}
\setlength{\tabcolsep}{6pt} 
\begin{tabular}{c c c c | c c c c}
\toprule
&\multicolumn{3}{c}{Single Camera} & \multicolumn{4}{c}{32-Camera Network}\\
&GPU & FPS & Power & Number & Number & FPS/ & Power \\
&  &   & [watts] & CPUs & GPUs & Camera & [watts] \\
\midrule
F. R-CNN & yes & 6.78 & 135 & -- & 32 & 6.78 & 4320 \\
YOLOv3 & yes & 14.79 & 153 & -- & 32 & 14.79 & 4895  \\
OpenPose & yes & 6.2 & 175 & -- & 32 & 6.2 & 5600 \\
tiny YOLO & no & 17.78 & 102 & 4 & 0 & 2.9 & 520 \\
\textbf{Activity} & \textbf{no} & \textbf{30$+$} & \textbf{50} & \textbf{1} & \textbf{0} & \textbf{25} & \textbf{80} \\
\end{tabular}
   \label{tbl:benchmarkTable}
\end{table}

Table \ref{tbl:benchmarkTable} shows in general that the pixel-based activity detector is much more computationally efficient than the object detection approaches. For a single camera, the activity detector can run at (and above) the rate of a 30 frames per second video feed, whereas the neural methods run at half or less rate. CPU power required for the activity detector is about half of Tiny Yolo on a CPU and about a third of the other object detectors running on a GPU. 

In an industrial application, for instance, where there are multiple cameras, the difference is more compelling as seen on the right side of Table \ref{tbl:benchmarkTable}. On our test machine, we can perform activity detection on up to 32 cameras. Using this as a baseline, we compare for a network of 32 cameras at which activity detection drops to 25 fps. For this workload, Tiny Yolo requires 4 CPUs and the frame rate drops to 6.2 fps. Frame rate for the other methods stays the same as for 1 camera because each of these uses a full GPU per single camera feed. Besides cost of GPUs, energy usage is an important system consideration for real applications. Power consumption is about $5 \times$ greater for Tiny Yolo than activity detection and over $50 \times$ greater for the other methods. These results support using activity detection alone or as a pre-filter for less frequent object detection as will be discussed in section \ref{sec:eventFltr}.

\subsection{Hybrid Activity Filter and Object Detector}
\label{sec:eventFltr}

The experimental results of Section \ref{sec:Comparing} show that performing activity detection is much more efficient than object detection on every frame. But, what if we want more detailed information than the presence or absence of activity? When this is the case, we can employ activity detection as a pre-filter (or gate) to perform or not perform object detection. The argument for a hybrid approach such as this is dependent upon the application and the activity density. For instance, if activity is constant, we might just as well perform object detection on all frames. If there are periods of inactivity, then use of the activity pre-filter is more efficient. 

We have an example of real data where activity in a business hallway was monitored for 2 years (activity plot shown in Fig. \ref{fig:activityPlots}). There were, on average, 300 activity events per camera per workday. If an event duration is 10 seconds, only $8.3\%$ of camera time contains an event. 

Using the 32-camera numbers from Table \ref{tbl:benchmarkTable}, if we performed YOLOv3 on one frame of each event detected by the activity detector, then the extra cost above activity detection is 1 GPU and 28w-h (watt-hours). This is the hybrid approach shown in Table \ref{tbl:eventCmpTable}. If we were to run continuous YOLOv3 neural network processing to do object detection without a pre-filter, this incurs an extra cost of 32 GPUs and $60\times$ the energy.

\begin{table}[t]
\centering
   \caption{Activity and object detection with an average of 300 events in a workday.}
\setlength{\tabcolsep}{6pt} 
\begin{tabular}{c c c c | c c c}
\toprule
&\multicolumn{3}{c}{Single Camera} & \multicolumn{3}{c}{32-Camera Network}\\
Detection&Number & Number & Energy & Number & Number & Energy \\
&CPUs & GPUs & [w-h] & CPUs & GPUs & [w-h] \\
\midrule
Activity& 1 & 0 & 500 & 1 & 0 & 800 \\
\textbf{Hybrid}& \textbf{1} & \textbf{1} & \textbf{500.9} & \textbf{1} & \textbf{1} & \textbf{828} \\
Object& 1 & 1 & 2030 & 1 & 32 & 49460 \\

\end{tabular}
   \label{tbl:eventCmpTable}
\end{table}

\subsection{Incorporation into ROS}
\label{sec:DeploymentROS}

\begin{figure}[t]
\centering
    \includegraphics[width=0.6\linewidth]{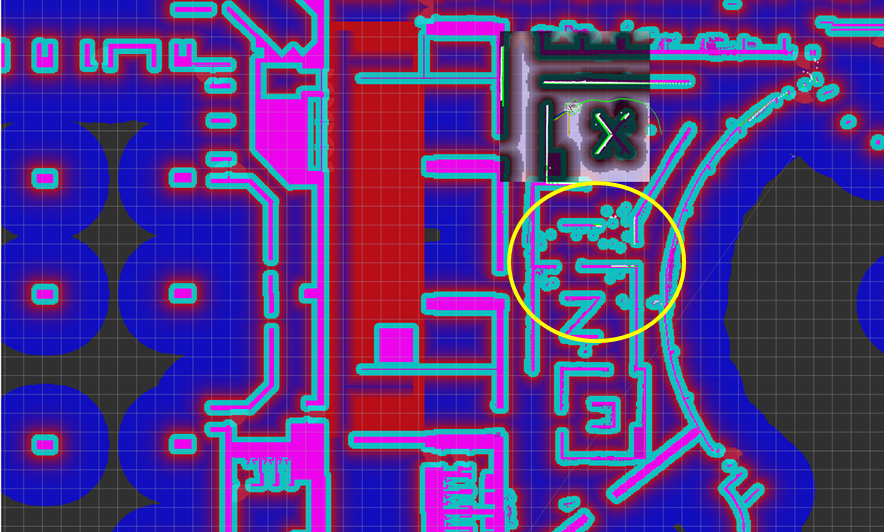}
   \caption{ROS cost map showing people activity with cyan dots inside yellow circle.}
   \label{fig:ROScostmap}
\end{figure}

In practice, we manage our robots on the Robot Operating System (ROS) \cite{ROS}. A preliminary task in using ROS is to populate a cost map with a floorplan of walls and other static objects. By assigning cost values to (x,y) locations, walls can be designated impenetrable, regions can be marked forbidden, and buffers zones can be placed around objects to help guide robots along safe and efficient paths. Fig. \ref{fig:ROScostmap} shows a ROS cost map of our robot test area. Walls are marked in pink with cyan buffer zones, red marks forbidden zones, and blue marks low-cost areas preferable for robot travel.

We augment the static cost map with dynamic human activity cost determined by temporal filtering. The yellow circle in Fig. \ref{fig:ROScostmap} indicates a region of human activity as shown by cyan dots. Just as the robot avoids cyan buffer zones, it will also avoid the cyan activity locations -- the difference being that the activity locations can move as they are detected in different locations. We currently assign the same cost to human activity as for static objects. However, it is reasonable, since humans can move, that a lower cost could be assigned to human activity, with the balance between safety and efficiency being a factor in choosing that value.

\section{Conclusions}
\label{sec:Conclusions}

Although mobile robots carry sensors to aid navigation, there are complementary benefits from fixed cameras that view the paths that robots travel. A major benefit is continuous view of an area from which a cycle of activity can be determined. Through knowledge of daily activity patterns, long-term path planning can be performed to avoid areas and times that are crowded, and instead choose paths at off-peak times. We have shown that a cascade filter applied to activity captured in both isochronal and chronological time can efficiently provide activity information for detecting long- and short-term activity. Furthermore, we have shown that a hybrid solution of temporal filtering for event detection, followed by object detection can yield power and cost efficiencies.


%
%
%
{\small
\bibliographystyle{splncs04}
\bibliography{bib}
}

\end{document}